\pdfoutput=1

\documentclass[runningheads,a4paper]{llncs}

\usepackage{amssymb}
\usepackage{graphicx}
\usepackage{times}
\usepackage{epsfig}
\usepackage{amsmath}
\usepackage{amssymb}
\usepackage{subfig}

\usepackage{url}
\newcommand{\keywords}[1]{\par\addvspace\baselineskip
\noindent\keywordname\enspace\ignorespaces#1}

\begin{document}

\mainmatter  

\title{A Semi-Automated Method for Object Segmentation in Infant's Egocentric Videos to Study Object Perception}

\titlerunning{Proposing a semi-automated method for object segmentation}

%
%
\author{Qazaleh Mirsharif\inst{1}
\and Sidharth Sadani\inst{2}\and Shishir Shah\inst{1}\and\\ Hanako Yoshida\inst{3}\and
Joseph Burling\inst{3} }
\authorrunning{Proposing a Semi-automated method for object segmentation in infant's egocentric videos}
\institute{Department of Computer Science, \\
University of Houston, Houston, TX 77204, USA,\\
\email{Qazaleh.mirsharif@gmail.com}\\
\email{Sshah@central.uh.edu}\\
\and Department of Electronics \& Communication, \\
Indian Institute of Technology, Roorkee, India,\\
\email{Sidharthsadani@gmail.com}\\
\and Department of Phsycology,\\
University of Houston, 126 Heyne Building Houston, TX 77204-5022 USA\\
\email{Yoshida,Jmburling@uh.edu}}

%

%
%
%
\toctitle{Proposing a semi-automated method for object segmentation}
\tocauthor{Authors' Instructions}
\maketitle

\begin{abstract}
Object segmentation in infant's egocentric videos is a fundamental step in studying how children perceive objects in early stages of development. From the computer vision perspective, object segmentation in such videos pose quite a few challenges because the child's view is unfocused, often with large head movements, effecting in sudden changes in the child's point of view which leads to frequent change in object properties such as size, shape and illumination. In this paper, we develop a semi-automated, domain specific, method to address these concerns and facilitate the object annotation process for cognitive scientists allowing them to select and monitor the object under segmentation. The method starts with an annotation from the user of the desired object and employs graph cut segmentation and optical flow computation to predict the object mask for subsequent video frames automatically. To maintain accuracy, we use domain specific heuristic rules to re-initialize the program with new user input whenever object properties change dramatically. The evaluations demonstrate the high speed and accuracy of the presented method for object segmentation in voluminous egocentric videos. We apply the proposed method to investigate potential patterns in object distribution in child's view at progressive ages. 
\end{abstract}

\section{Introduction}

Infants begin to learn about objects, actions, people and language through many forms of social interaction. Recent cognitive research highlights the importance of studying the infant's visual experiences in understanding early cognitive development and object name learning ~\cite{pereira_bottom-up_2014,pereira_first-person_2009,smith_not_2011}. The Infant's visual field is dynamic and characterized by large eye movements and head turns owing to motor development and bodily instabilities which frequently change the properties of their visual input and experiences. What infants attend to and how their visual focus on objects is structured and stabilized during early stages of development has been studied to understand the underlying mechanism of object name learning and language development in early growth stages \cite{bambach_understanding_2013,xu_its_2011,pereira_bottom-up_2014,burling_significance_2013}. 
 
Technological advancement allows researchers to have access to these visual experiences that are critical to understanding the infant's learning process first hand. ~\cite {yoshida_whats_2008,pereira_bottom-up_2014}. Head cameras attached to the child's forehead enables researchers to observe the world through child's viewpoint by recording their visual input ~\cite{smith_contributions_2014,pereira_first-person_2009}. However, it becomes very time consuming and impractical for humans to annotate objects in these high volume egocentric videos manually. 

As discussed in ~\cite {bambach_survey_2015}, egocentric video is an emerging source of data and information, the processing of which poses many challenges from a computer vision perspective. Recently, computer vision algorithms have been proposed to solve the object segmentation problem in such videos ~\cite {ren2010figure,ren2009egocentric}. The nuances of segmentation in egocentric videos arise because the child's view is unfocused and dynamic. Specifically, the egocentric camera, (here, the head camera) is in constant motion, rendering the relative motion between object and background more spurious than that from a fixed camera. In addition, the random focus of a child causes the objects to constantly move in and out of the view and appear in different sizes, and often the child's hand may occlude the object. Examples of such views are shown in figure 1 (a,b and c). Finally, if the child looks towards a light source, the illumination of the entire scene appears very different, as shown in figure 1(d).


In this paper, we develop an interactive and easy to use tool for segmentation of objects in child's egocentric video that addresses the above problems. The method enables cognitive scientists to select the desired object and monitor the segmentation process. The proposed approach exploits graph cut segmentation to model object and background and calculate optical flow between frames to predict object mask in following frames. We also incorporate domain specific heuristic rules to maintain high accuracy when object properties change dramatically.
 
The method is applied to find binary masks of objects in videos collected by placing a small head camera on the child as the child engages in toy play with a parent. The object masks are then used to study object distribution in child's view at progressive ages by generating heat maps of objects for multiple children. We investigate the potential developmental changes in children's visual focus on objects. 

 The rest of the paper is organized as follows: We describe the experimental setup and data collection process in section 2. The semi automated segmentation method is explained in detail in section 3. Results are presented and discussed in section 4. Finally section 5 will conclude the present study and highlights the main achievements and contributions of the paper.

\section{The Experimental Setup}

A common approach to study infant's cognitive developmental process is to investigate their visual experience in a parent-child toy play experiment. In the current study we extract the infant's perspective by placing a head camera on the child and recording the scene as it engages in tabletop toy play sitting across from the mother. Each play session is around 5 minutes where the mother plays with toys of different colors and sizes including bunny, carrot, cup, cookie and car. A transparent jar was initially among the toys, but was removed from the study as the proposed method does segment such objects accurately. The mother plays with the toys one by one and names the toys as she attempts to bring the infant's attention to that toy. Multiple toys may sit in the view at the same time. The videos are recorded at progressive ages including 6,9,12,15 and 18 months. In this paper we have used 15 videos which consist of three videos from each age.  From each video, we extracted approximately 9500 image frames. The image resolution is 480 to 640 pixels. 	

\section{The Proposed Approach}
In this section we explain our proposed method in detail in three main steps namely, initialization and modeling of the object and background, object mask prediction for next frame, and performing a confidence test to continue or restart the program. The flow diagram for the method is shown in figure 2. We use a graph based segmentation approach \cite{boykov_fast_2001} \cite{boykov_experimental_2004} to take user input, for initialization and also when recommended by the confidence mechanism (Sec 3.3). For each user input we model the object and save the features as a KeyFrame in an active learning pool, which is used as ground truth data. We then use optical flow \cite{horn_determining_1981} to estimate the segmentation mask for the next frame and subsequently refine it to obtain the final prediction mask (Sec 3.2). The obtained segmentation result is then evaluated under our confidence test (Sec 3.3). If the predicted mask is accepted by the test, the system continues this process of automatic segmentation. When the confidence test fails, the system quickly goes back and takes a new user input to maintain the desired accuracy in segmentation. 

\begin{figure}[t,h]
	\begin{center}
		\subfloat[]{
			\includegraphics[width=0.77 in]{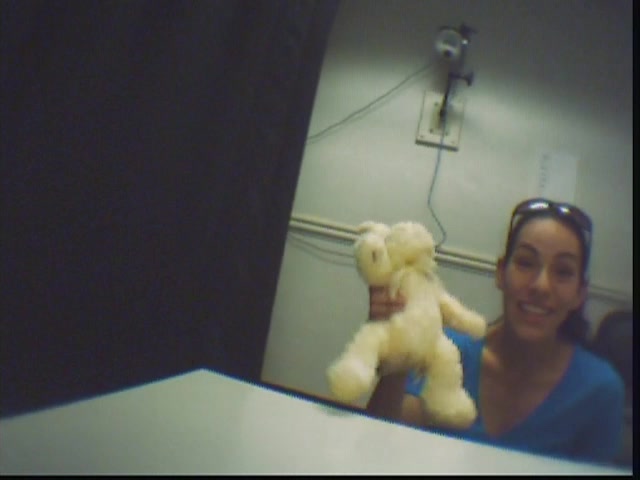}
			\includegraphics[width=0.77 in]{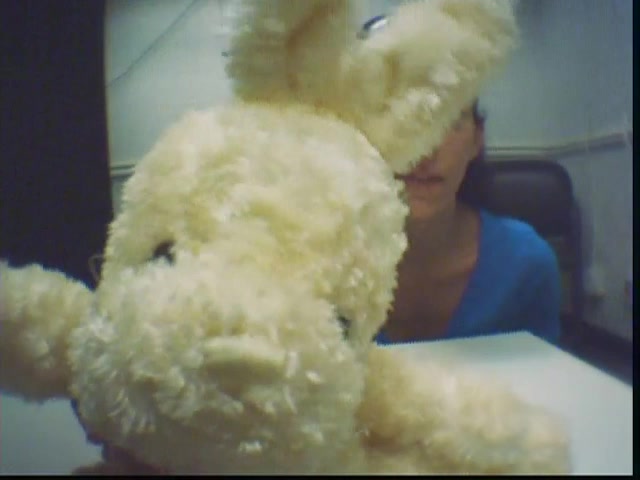}}
		\subfloat[]{
			\includegraphics[width=0.77 in]{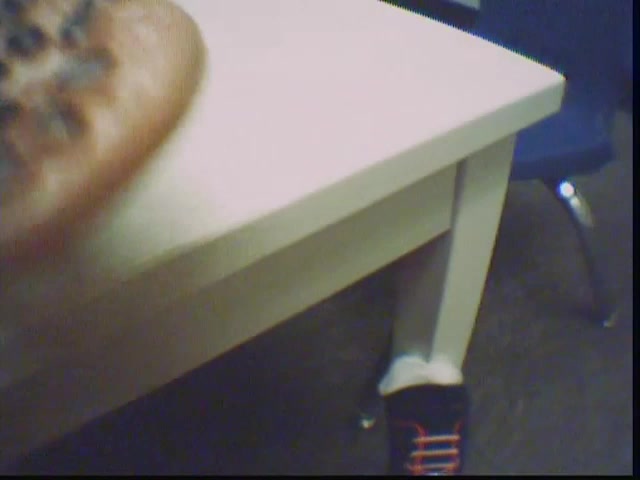}}
		\subfloat[]{
			\includegraphics[width=0.77 in]{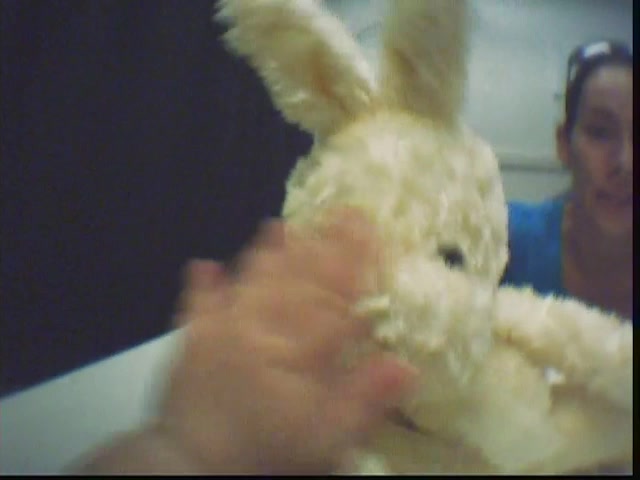}}
		\subfloat[]{
			\includegraphics[width=0.77 in]{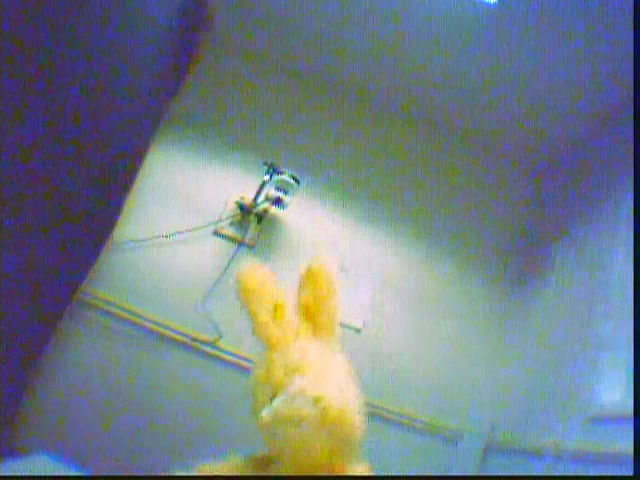}}
		\subfloat[]{
			\includegraphics[width=0.77 in]{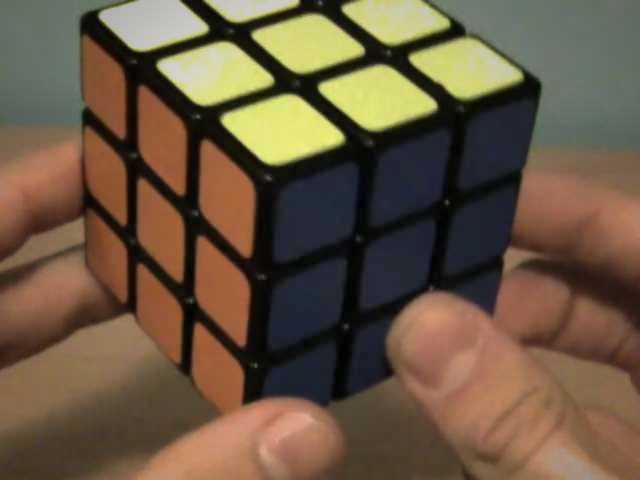}}
		
	\end{center}
	\caption{(a) Size Variation (b) Entering and Leaving (c) Occlusion (d) Illumination Variation (e) Orientation Variations}
	\label{fig:Challenges}
\end{figure}

\begin{figure}[t,h]
	\begin{center}
		
		\subfloat[]{
			\includegraphics[width=4 in]{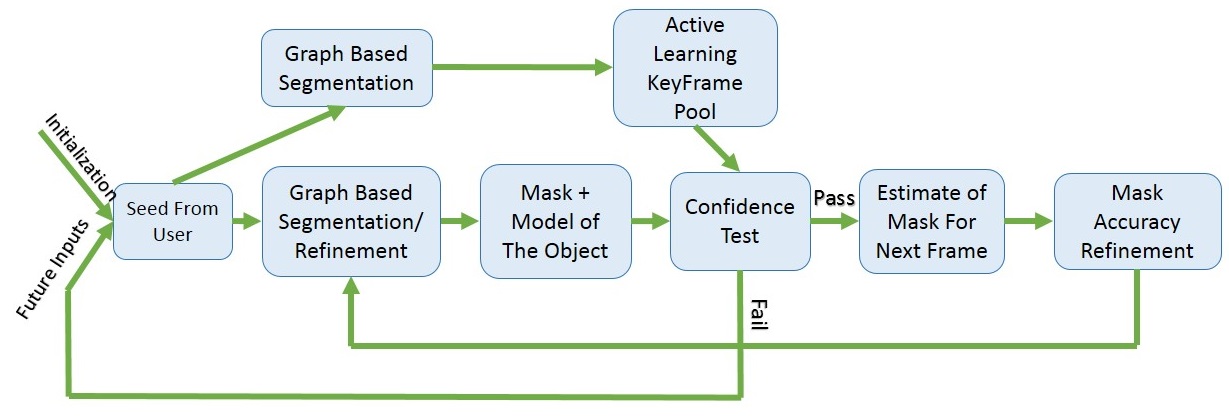}}
		
	\end{center}
	\caption{Flow diagram of the proposed approach }
	\label{fig:Flow}
\end{figure}
 

\subsection {Initialization, User Input and Object Model} 
We initialize the segmentation algorithm with a user seed manually encompassing the object in a simple polygon. We then use an iterative Graph-Cut \cite{boykov_experimental_2004} based segmentation approach to segment the desired object. Given that we work with opaque objects this method suits well to our requirement of accuracy. We calculate the minimum cut of the graph whose nodes are each of the pixels. Each of the nodes (pixels) are given a foreground $f_{i}$ and background prior $b_{i}$. For the initial frame this prior is calculated from the user seed, and for each subsequent frame the prior is calculated from the model of the foreground and background of the previous frame. In our approach we use a Multivariate Gaussian Mixture Model (GMM) of the RGB components in the image, as our model for the object as well as the background. 

Along with the prior term we use a normalized smoothness cost term $S_{ij}$, which is a penalty term if two adjacent pixels have different assignments. As mentioned in \cite{boykov_experimental_2004}, a normalized gradient magnitude obtained from the sobel operator is used for the same. We run the Graph-Cut algorithm \cite{boykov_experimental_2004} iteratively until the number of changes in the pixel assignments between two consecutive iterations falls below a certain acceptable threshold. The Multivariate GMM is updated in each iteration using kmeans clustering. The mask along with the Multivariate GMM define the combined model of the object and are the starting point of the prediction of the mask for the next frame. 

\subsection{Segmentation Prediction for Next Frame}
We begin our prediction with the calculation of dense optical flow between the previous frame and the current frame. Since the two frames are consecutive frames of an egocentric video, we can assume there isn't a drastic change in the characteristics of the foreground or the background. Using optical flow we predict pixel to pixel translation of the mask. This initial calculation provides a starting estimate of the mask. 

\begin{multline}
(x,y)_{CurrentMask} = (x,y)_{PreviousMask} + v(x,y)
\\ (x,y) \in Pixel \ Coordinates \ of \ Foreground \ , 
\\ v(x,y) = Optical \ Flow \ of \ Pixel (x,y) 
\end{multline}

Some refinement in this mask is required to maintain the accuracy for the following reasons. Firstly, the pixel to pixel transformation using optical flow is not a one to one but a many to one transformation i.e. many pixels in the previous frame may get translated to the same pixel in the current frame. Secondly, if part of the object is just entering the frame from one of the boundaries, optical flow by itself cannot predict if the boundary pixels belongs to the object or not. Lastly, under occlusion, as is the case in many frames when the mother is holding the object or the child's hands are interacting with the object, flow estimates the mask fairly well at the onset of occlusion but fails to recover the object once the occlusion has subsided thus leading to severe under segmentation. 

\begin{figure}[h]
	\begin{center}
		\subfloat[]{\includegraphics[width=0.83 in]{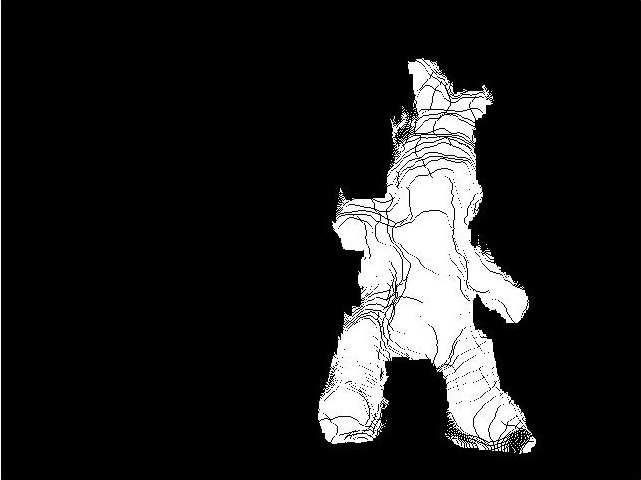}}
		\subfloat[]{\includegraphics[width=0.83 in]{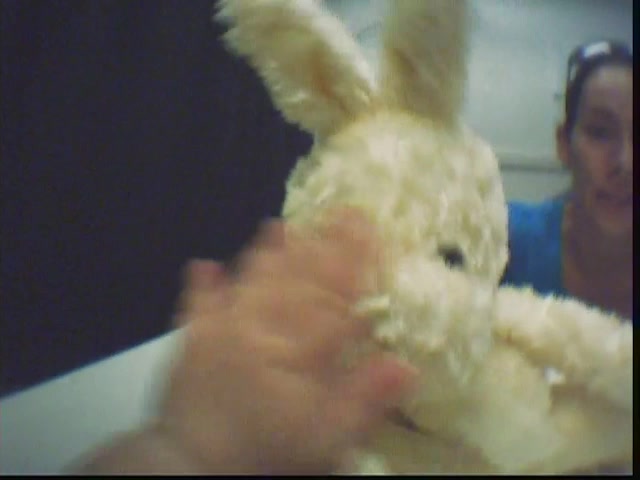}}
		\subfloat[]{\includegraphics[width=0.83 in]{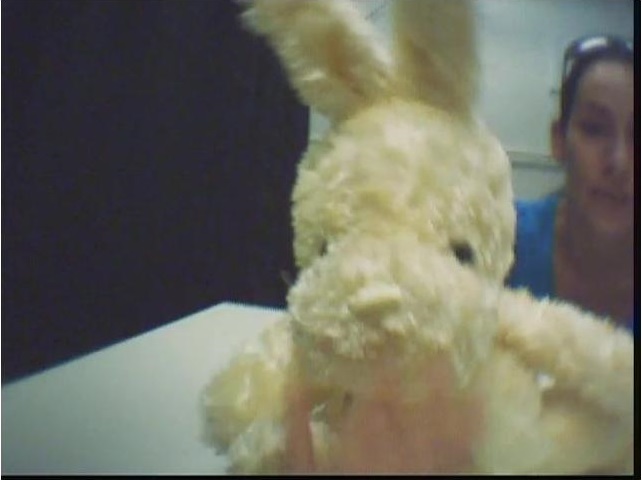}}
		\subfloat[]{\includegraphics[width=0.83 in]{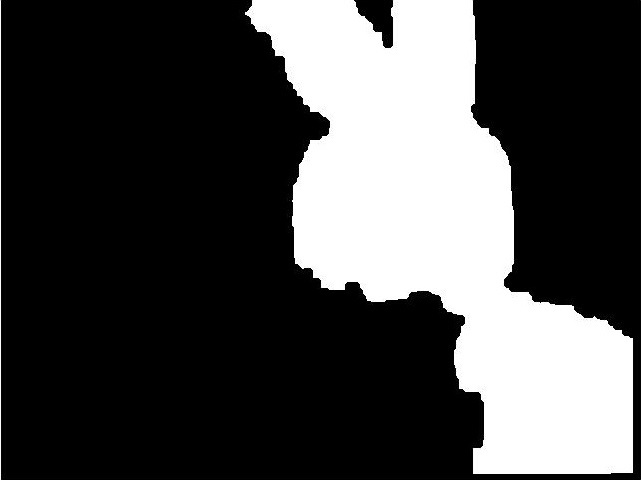}}
	\end{center}
	
	\caption{Need For Refinement (a) Error due to Optical Flow (b)-(c) Occlusion of object by the child's hand in successive frames (d) Predicted Mask }
	\label{fig:Errors}
\end{figure} 

To refine this initial estimate of the mask, we define a region of uncertainty around this initial estimate of the mask, both inwards and outwards from the mask. If the object happens to be near one of the edges of the frame, we define areas along the right, left, top, or bottom edges as part of the uncertain region based on the average flow of the object as well as the local spatial distribution of the object near the edges. We then input this unlabeled, uncertain region into the earlier Graph-Cut stage to label these uncertain pixels as either foreground or background. This helps in obtaining a more accurate, refined segmentation mask.

\begin{figure}[h]
	\begin{center}
		\subfloat[]{\includegraphics[width=1 in]{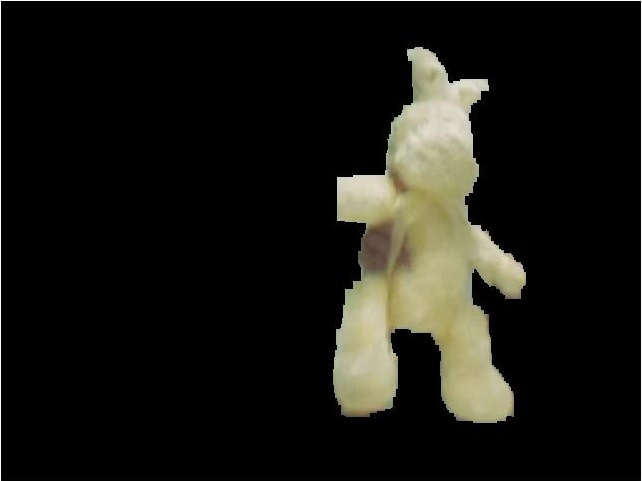}}
		\subfloat[]{\includegraphics[width=1 in]{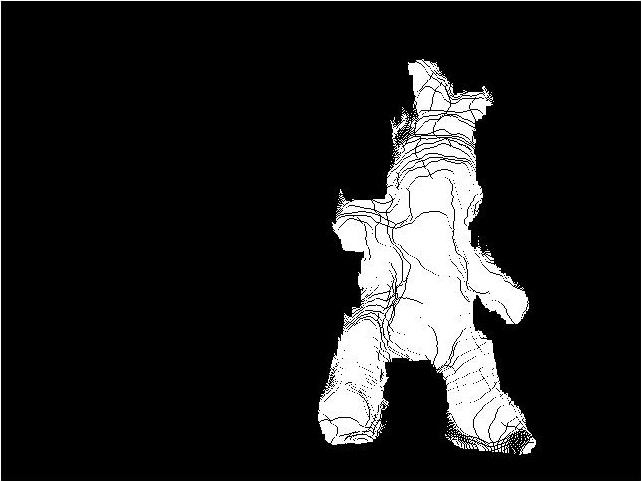}}
		\subfloat[]{\includegraphics[width=1 in]{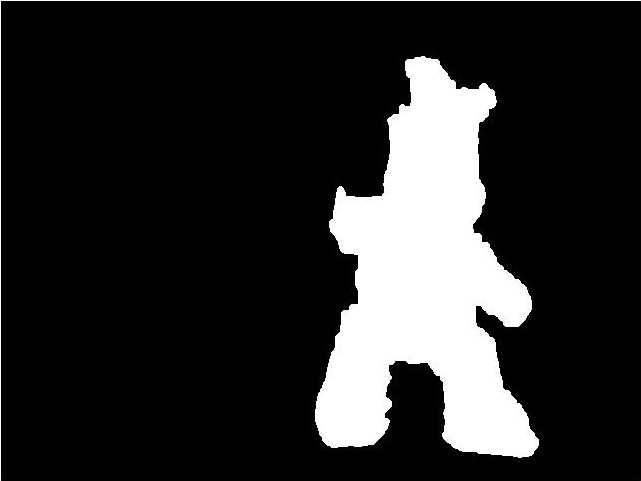}}
		\subfloat[]{\includegraphics[width=1 in]{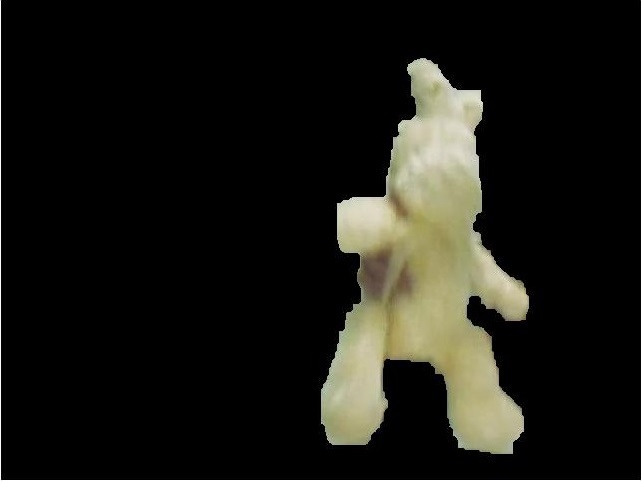}}
	\end{center}
	
	\caption{Steps in Segmentation Prediction (a) User Seed (b) First estimate of next mask using optical flow (c) The Uncertain Region (d) Final mask using Graph Cut (Bunny moved downwards, legs expanded) }
	\label{fig:Masks}
\end{figure} 

This segmentation result is now compared against the learnt ground truth from the user, stored in the active learning KeyFrame Pool based on a confidence test (explained in the next section) . If the segmentation result is accepted by the confidence test, we go on to predict the mask for the next frame using this mask. If the result fails the confidence test, we go back and take a new input from the user.

\subsection{Confidence Test, Learning Pool, Keyframe Structure}
We define Keyframes as those frames in which the segmentation mask has been obtained from user input. This represents the ground truth data. For each keyframe we save the following parameters :

\begin{itemize}
	\item Size of The Segmentation Mask in Pixels
	\item Multivariate GMM parameters for foreground and background (Centers, Covariances, Weights)
	\item The amount of acceptable error in the centers of the Modes of the GMM
\end{itemize}

These three parameters are utilized to test the validity of the predicted segmentation mask. Firstly, we check if the size of the mask has increased or decreased beyond a certain fractional threshold of the size of the mask in the most recent keyframe. This test is introduced as a safeguard to maintain the reliability of segmentation because when the child interacts with the object, brings it closer or moves it away, the object may suddenly go from occupying the entire frame to just a small no. of pixels. We've implemented to flag 39$\%$ variation, but anywhere between 25$\%$-50$\%$ can be chosen depending on the sensitivity of the results required.

\begin{multline}
Size_{CurrentFrame} \leq 0.61*Size_{KeyFrame} \quad (or) 
\\ Size_{CurrentFrame} \geq 1.39*Size_{KeyFrame} \ ; Confidence = False
\end{multline}

Secondly, we look to flag under segmentation, mostly during recovery of the entire segment after occlusion. To detect under segmentation we assume that the background cluster of the current frame moves closer or is similar to the foreground cluster of the most recent keyframes to account for the presence of the incorrectly labelled foreground pixels as background. We hypothesize that the background cluster that moves closer to the foreground does so very slightly as it still must account for the background pixels but the length of the projection of the eigen vectors along the line joining the centers of this background - foreground cluster pair increases. 

\begin{equation}
\begin{split}
D_{BFiKey}=\min\limits_{j}||\vec {BC_{i,Key}}-\vec{FC_{j,Key}}||^2  \\
D_{BFiCurr}=\min\limits_{j}||\vec {BC_{i,Curr}}-\vec{FC_{j,Curr}}||^2 \\
\end{split}
\end{equation}
From the above two distances, the first one being for the keyframe clusters and the second for the current frame, we can find which background cluster moved closer to which foreground cluster. After which we calculate the projections of the eigen vectors of the Background Cluster that moved closest to the Foreground Cluster onto the line joining the centers of these two clusters, in the keyframe and the current frame.

\begin{equation}
\begin{split}
P_{Key}=\sum\limits_{k=1}^{3}||\vec {E.Vec_{k,Key}}\cdot\vec{CC_{j,Key}}|| \ ,  
P_{Curr}=\sum\limits_{k=1}^{3}||\vec {E.Vec_{k,Curr}}\cdot\vec{CC_{j,Curr}}||
\end{split}
\end{equation}

If this increase is greater than 25$\%$ then we can reliably conclude under segmentation.

Lastly, the object may appear differently in different orientations or in different illumination conditions, for which we compare the GMM for the predicted segment against all the Models in the Keyframe. We do this by checking if the average distance between corresponding centers in the two segments is within the acceptable error for that GMM, and if so, are the difference in weights of these corresponding centers also within a certain acceptable threshold. The latter threshold is set manually depending on the number of modes and the desired sensitivity. The former is obtained using standard deviation of the RGB channels

\begin{equation}
\begin{split}
Thresh \ For \ Avg \ Dist \ For \ Center \ i =(\sum\limits_{k=1}^{3}Std_{i,k})/3 \\
where \quad k \in (R,G,B \ the \ 3 \ Channels)
\end{split}
\end{equation}

If this criteria is not met then we take a new user input and it becomes a keyframe in the learning pool.

%

\section{Results and Discussions}

We use the method to extract multiple objects from videos and compare the resulting object masks with their corresponding manual annotation provided by experts. Further, the performance of the algorithm in terms of run time and amount of user interaction, for each of these objects is reported in table 1. The run time of the method consists of the time taken for optical flow calculation and the processing time required by the proposed method. We see clearly that over a large number of frames the average total time would easily outperform the time required in manual segmentation. We also observe that the processing time varies directly with the (image size) area occupied by the object. Lastly we observe how frequently the algorithm requires user input. We let the method run for a large number of frames (300 frames, a subset of the entire video) and count the number of of user input requests. It is important to note that we have set the method to take user input every 50 frames, so even in case of no errors we would take 6 user inputs. Hence the additional user inputs required, due to uncertainty in prediction, are 5,3,3,4,7 on average, respectively. In any case this is a significant reduction in the amount of user involvement as only 3$\%$ of the frames require user input on average.
 
\begin{table}[b]
\centering
\begin{tabular}{||c c c c c c c c||} 
 \hline
 Object & Total & Optical Flow & Processing &  Image  & $\%$ Area & User Input & Accuracy\\
  & Time(sec) & Time(sec) & Time(sec)  & Size(px) &   & (per 300 frames) \\ [0.5ex] 
 \hline\hline
Bunny & 8.7002 & 8.0035 & 0.6967 & 199860 & 65.00$\%$ & 11 & 97.76$\%$ \\ 
Cup & 8.1156 & 7.6534 & 0.4622 & 47261 & 15.38$\%$ & 9 & 98.43$\%$\\ 
Carrot & 7.9513 & 7.5180 & 0.4333 & 32080 & 10.44$\%$ & 9 & 99.71 $\%$ \\
Car & 7.9237 & 7.6320 & 0.2917 & 10872 & 3.54$\%$ & 10 & 99.35$\%$\\
Cookie & 7.9421 & 7.6145 & 0.3276 & 27653 & 9.00$\%$ & 13 & 98.10 $\%$\\ [0ex] 
 \hline
\end{tabular}
\caption{Performance Measures (Note : The above values are average over 300 frames)}
\label{table:4}
\end{table}
 
As with any automated approach to segmentation, user interaction and processing time is reduced, what is traded off is the accuracy of the automated segmentation as compared to manual segmentation. To evaluate this we take 30 randomly picked frames and have them manually annotated by 5 different people. We calculate the DICE similarity between automated masks vs. the manually annotated masks and then the DICE similarity between the manually segmented masks for each of the frames for each pair of people. The mean and standard deviations of which are noted in table 2. We see that, on average, we lose only 3.21 percent accuracy as compared to manual segmentation.   Note: DICE similarity is measured as the raio of twice the no. of overlap pixels to the sum of the no. of pixels in each mask.

In our application, under segmentation is not tolerable but slight over segmentation is. We see that our approach doesn't undersegment any worse than manual segmentation would, as can be seen from the recall values in the two columns. On the other hand, our algorithm consistently, but not excessively (as we see from the DICE measurements), oversegments the object, as can be seen from the precision values in the two columns. Thus we see that the proposed approach significantly reduces time and user interaction with little loss in accuracy as compared to manual segmentation. Note : Precision is the proportion of mask pixels that overlap with manual segmentation and Recall is the proportion of the manual segmentation pixels that are part of the predicted mask.

\begin{table}[t]
\centering
\begin{tabular}{||c c c c ||} 
 \hline
 'DICE' & Our Algorithm vs. Manual & & Manual vs. Manual \\ [0ex] 
 \hline
 Mean & 0.9248 & & 0.9569  \\ 
 Std & 0.043 & & 0.0167  \\ [0ex] 
 \hline \hline
 'Precision \& Recall' & Our Algorithm vs. Manual & & Manual vs. Manual \\ [0ex]
\hline
 Precision & 0.8746 & & 0.9532  \\ 
 Recall & 0.9567 & & 0.9734  \\ [0ex]
\hline
\end{tabular}
\caption{DICE Similarity Coefficient \& Precision \& Recall Values}
\label{table:5}
\end{table}

We use the results obtained from segmentation to investigate how objects are distributed in the child's view at progressive ages. We look to study potential regularities in object movement patterns and concentration in child's view with age. To visualize the areas where infants focus and fixate on objects we plot the heat maps of object for each video. To see which locations have been occupied by objects most recently, we give each pixel of the object a weight $W_{i}$ for each object mask and accumulate the results in time. The final output stores the following values for each image pixels:


\begin{equation}
\begin{split}
P_{xy, Output}= \sum_{i=1}^{L} W_{i}*P_{xy,ObjectMask}\ ,  
W_{i}= i/L
\end{split}
\end{equation}

Where $i$ is the frame number and $L$ is the total number of frames in video (usually around 9500). 

From the heat maps, we can see that object movement in 6 months infants is large and does not follow any specific pattern. The object concentration region changes across infants and their visual focus on objects are not stabilized. However after 9 months, the object distribution pattern becomes more structured and the object concentration area moves down toward the middle-bottom part of the visual field. This region seems to be the active region where child interacts with object most of the time. For 18 months old children, object movements increase and the pattern change across the children. However the concentration point is still in the bottom area very close to child's eyes. The results might be aligned with previous psychological hypothesis which discusses an unfocused view in 6 month old infants and increasing participation of child in shaping his visual field with age ~\cite{yoshida_dynamic_2013}. 18 month old infants are able to make controlled moves and handle objects. This study is still at early stages and more investigation of other factors such as who is holding the object is required to discover how child's visual focus of attention is shaped and stabilized over the early developmental stages and who is shaping the view at each stage. The results demonstrate a developmental trend in child's visual focus with  physical and motor development which might support a controversial psychological hypothesis on existence of a link between physical constraint and language delay in children suffering from autism.

\begin{figure}[t]
	\begin{center}
		\subfloat[]{
			\includegraphics[width=0.8in]{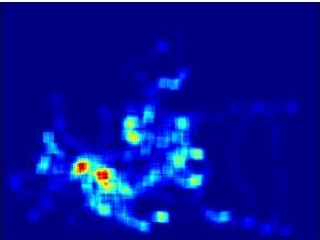}}
		\subfloat[]{
			\includegraphics[width=0.8in]{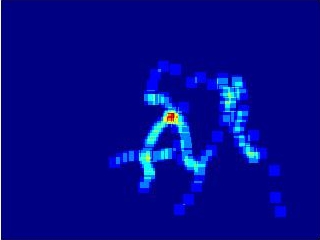}}
		\subfloat[]{
			\includegraphics[width=0.8in]{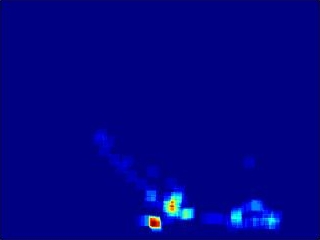}}
		\subfloat[]{
			\includegraphics[width=0.8in]{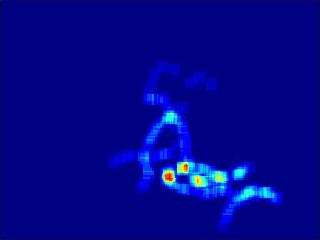}}
		\subfloat[]{
			\includegraphics[width=0.8in]{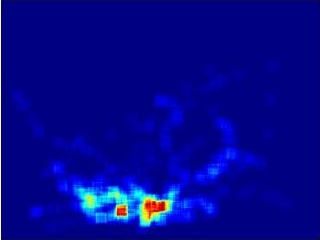}}\\
		\subfloat[]{
			\includegraphics[width=0.8in]{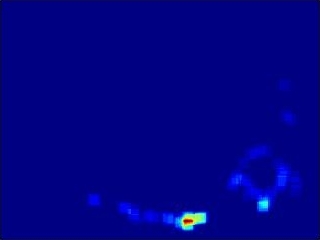}}
			\subfloat[]{
			\includegraphics[width=0.8in]{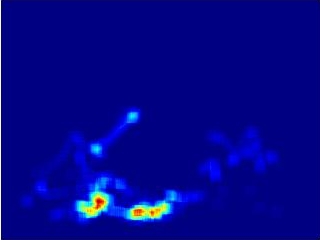}}
		\subfloat[]{
			\includegraphics[width=0.8in]{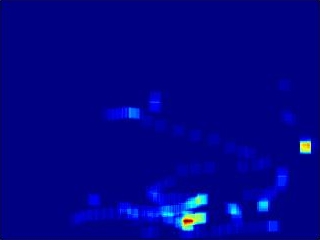}}
			\subfloat[]{
			\includegraphics[width=0.8in]{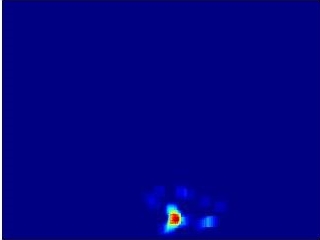}}
		\subfloat[]{
			\includegraphics[width=0.8in]{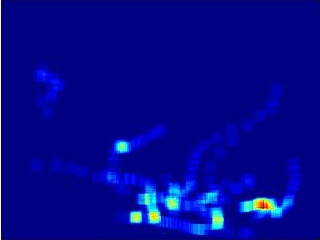}}	
						\end{center}
	
	\caption{Heatmaps of objects for infants at progressive ages (a,b) 6 months infants (c,d) 9 months infants (e,f) 12 months infants  (g,h) 15 months infants(i,j) 18 months infants }
	\label{fig:teaser}
\end{figure} 
\section{Conclusions}

We proposed a semi-automated method for object segmentation in egocentric videos. The proposed method uses domain specific rules to address challenges specific to egocentric videos as well as object occlusion and allows researchers to select the desired object in the video and control the segmentation process. The method dramatically speeds up the object annotation process in cognitive studies and maintain a high accuracy close to human annotation. 
We applied the method to find object masks for a large number of frames and studied object movement and concentration patterns in child's visual field at progressive ages. We found a developmental trend in child's visual focus of attention with age and movement of active region toward middle bottom region of child's visual field. The current study is one step toward understanding the mechanism behind early object name learning and cognitive development in humans.

\bibliographystyle{splncs03}
\bibliography{typeinst}


\end{document}